\documentclass[10pt, letterpaper]{article}
\usepackage[utf8]{inputenc}
\usepackage{siunitx}
\usepackage{graphicx}
\usepackage{subcaption}
\graphicspath{{./} }
\usepackage[style=nature]{biblatex}
\usepackage{textgreek}
\usepackage[margin=25mm]{geometry}
\usepackage{geometry}
\usepackage{amsmath}
\usepackage{array}
\usepackage{comment}
\usepackage[table]{xcolor}
\usepackage{color, colortbl}
\definecolor{MyColor1}{rgb}{0.8,1,1}
\definecolor{MyColor2}{rgb}{1,1,0.7}

\begin{document}

\title{High Resolution Solar Image Generation using Generative Adversarial Networks}

\author{Ankan Dash, Junyi Ye, Guiling Wang \\
        \small Department of Computer Science \\ \small New Jersey Institute of Technology, Newark, NJ 07102, United States \\
}

\date{}
\maketitle

\begin{abstract}
We applied Deep Learning algorithm known as Generative Adversarial Networks (GANs) to perform solar image-to-image translation. That is, from Solar Dynamics Observatory (SDO)/Helioseismic and Magnetic Imager(HMI) line of sight magnetogram images to SDO/Atmospheric Imaging Assembly(AIA) 0304-Å images. The Ultraviolet(UV)/Extreme Ultraviolet(EUV) observations like the SDO/AIA0304-Å images were only made available to scientists in the late 1990s even though the magenetic field observations like the SDO/HMI have been available since the 1970s. Therefore by leveraging Deep Learning algorithms like GANs we can give scientists access to complete datasets for analysis. For generating high resolution solar images we use the Pix2PixHD and Pix2Pix algorithms. The Pix2PixHD algorithm was specifically designed for high resolution image generation tasks, and the Pix2Pix algorithm is by far the most widely used image to image translation algorithm. For training and testing we used the data for the year 2012, 2013 and 2014. The results show that our deep learning models are capable of generating high resolution(1024 x 1024 pixels) AIA0304 images from HMI magnetograms. Specifically, the pixel-to-pixel Pearson Correlation Coefficient of the images generated by Pix2PixHD and original images is as high as 0.99. The number is 0.962 if Pix2Pix is used to generate images. The results we get for our Pix2PixHD model is better than the results obtained by previous works done by others to generate AIA0304 images. Thus, we can use these models to generate AIA0304 images when the AIA0304 data is not available which can be used for understanding space weather and giving researchers the capability to predict solar events such as Solar Flares and Coronal Mass Ejections. As far as we know, our work is the first attempt to leverage Pix2PixHD algorithm for SDO/HMI to SDO/AIA0304 image-to-image translation.
\end{abstract}

\providecommand{\keywords}[1]
{
  \small	
  \textbf{\textit{Keywords:}} #1
}

\section{Introduction}
The sun is the main source of space weather due to its constant emission of energy and electromagnetic radiation in the form of light and electrically charged particles\cite{spaceWeather}. Several solar phenomena like Solar Flares, Coronal Mass Ejections (CMEs), High Speed solar winds and Solar energetic particles constitute the solar weather and cause space weather effects here on Earth\cite{ECHER2005855}. Solar Flares are giant bursts of energy and radiations comprising of X-Rays and Gamma Rays due to the sudden release of magnetic energy that has built up in the solar atmosphere\cite{Fletcher_2011}. Coronal Mass Ejections(CME) are giant clouds of magnetic plasma in the Sun’s corona that are hurled into space. CMEs can generate strong shock-waves that can accelerate solar wind and extend billions of miles into space\cite{CME}. The regions of the Sun known as coronal holes produce high-speed solar wind streams\cite{Garton_2018}. Solar energy particles are high-energy particles from the Sun that can come either from the site of a solar flare or from shock waves associated with coronal mass ejections\cite{reames2020solar}. It is extremely important to have the capabilities to predict the above solar events because astronauts run the risk of being exposed to solar radiation. Also, these solar events can interfere with proper functioning of satellites and cause power grid disruption which will be a major problem in today's digitally connected world\cite{marov}.

To give scientists and researchers prediction capabilities, NASA and several other Space organizations across the globe are constantly monitoring the Sun’s activities. Numerous space missions like the Solar Heliospheric Observatory (SOHO) \cite{Domingo1995}, Solar Terrestrial Relations Observatory (STEREO) \cite{Kaiser2008} and Solar Dynamics Observatory (SDO) \cite{Pesnell2012} capture various observations of the Sun at different wavelengths which include the Ultraviolet (UV), Extreme Ultraviolet (EUV) images, solar magnetograms and dopplergrams. The NASA Living With a Star (LWS)  \cite{BREWER2002609} is a program whose primary objective is to study the Sun and its effects on the Earth. The Solar Dynamics Observatory (SDO) is the first mission launched by NASA under the LWS program. The SDO takes high resolution images of the Sun at various wavelengths to help solar physicists understand the generation and the structure of the Sun’s magnetic field, and to have predictive capability of events like Solar Flares and Coronal Mass Ejections (CMEs). The SDO contains a variety of instruments like HMI (Helioseismic and Magnetic Imager) \cite{Scherrer2012}, AIA (Atmospheric Imaging Assembly) \cite{Lemen2012} and EVE (Extreme Ultraviolet Variability Experiment) \cite{Woods2012} to capture various observations of the Sun. The HMI instrument is led by Stanford University and focuses on the movement and magnetic properties of the sun's surface and provides 3 images. Managed by the Lockheed Martin Solar and Astrophysics Laboratory (LMSAL), the AIA instrument takes continuous full-disk observations of the solar chromosphere and corona at 10 different wavelengths measured in angstroms (Å). The EVE instrument is led by University of Colorado and measures the EUV irradiance.   

Deep Learning \cite{LeCun2015} has been revolutionizing many fields, from self-driving cars to medicine and it was only a matter of time before it found its way into physical sciences. Generative Adversarial Networks \cite{GoodfellowNIPS2014} or GANs for short are a type of generative models that were invented by Ian Goodfellow and his colleagues. Generative Models \cite{GM} try to learn a probability density function from a training set and then generate new samples that are drawn from the same distribution. Some of these models include Boltzmann Machines\cite{Hinton2010BoltzmannMachines}, Deep Belief Networks\cite{Hinton2010DeepBeliefNets} and Variational Autoencoders\cite{kingma2014autoencoding}. Generative Adversarial Networks\cite{GoodfellowNIPS2014} are by far the most widely used generative models. GANs work by pitting two neural networks (the Generator and the Discriminator) against each other to generate new synthetic images that are like the real images. Over the past couple of years, GANs have made significant progress. With the increase in model capacity, we are now able to train deeper and more complex Generator and Discriminator neural network architectures due to hardware improvements. GAN variations such as Progressive GAN \cite{karras2018progressive} enable us to generate high resolution images with improved training stability. Paired image-to-image translation GANs (Pix2Pix\cite{Isola}) and unpaired image-to-image translation GANs (CycleGAN \cite{Zhu}) can find a mapping between paired and unpaired pairs of images to solve the image-to-image translation problems. Even though Pix2Pix has been widely used for image translation(\cite{Kim2019}, \cite{Park_2019}) tasks, it is not well suited for the generation of high resolution images and it fails to capture finer details and can sometime cause artifacts in high resolution image to image translation tasks. To overcome this problem the Pix2PixHD algorithm proposed by Wang et al.\cite{WangPix2PixHD} can be used to generate high-res images with minimum artifacts and capture finer details of the image. 

\subsection{Motivation}
The UV/EUV observations like the ones taken by the SDO/AIA instrument were only made available after the launch of the SOHO observatory in the late 1990s. However, the magnetic field observations like the ones captured by the SDO/HMI instrument have been available since the 1970s.  UV and EUV observations which include the AIA0304 measurements help researchers to understand different areas of the sun, such as coronal loops, filaments, coronal holes, active regions \cite{DelZanna2018} and are thus vital in space weather prediction. 

Even though there have been some attempts for solar image generation using deep learning(\cite{Kim2019}, \cite{Park_2019}, \cite{Galvez_2019}, \cite{Shin_2020}, \cite{Jeong_2020}), most of them have focussed on the generation of Solar Farside Magnetograms or Magnetic Field observations from UV and EUV data (\cite{Kim2019}, \cite{Shin_2020}, \cite{Jeong_2020}) but very few have tried to generate UV and EUV images using magnetograms(\cite{Park_2019}). Also most of the works related to the generation of UV and EUV images from magnetograms have been using the Pix2Pix algorithm(\cite{Park_2019}).

In this work we implemented conditional Generative Adversarial Networks, to perform image-to-image translation. More specifically, we use the Pix2PixHD and the Pix2Pix models to generate high resolution SDO/AIA0304 images from SDO/HMI line of sight magnetograms. 

Here we have considered only the AIA-304 Å passband out of the 9 passband (94, 131, 171, 193, 211, 304, 335, 1600, and 1700 Å). This is because the magnetic fields themselves are invisible but as the charged plasma moves along the magnetic field lines these magnetic fields become visible at the extreme ultraviolet wavelength of 304 Å, showing material at a temperature of approximately 50000 Kelvin. This allows scientists to observe and analyze the magnetic fields on the sun. Even though we have considered only AIA-304 Å, this framework is compatible with other passbands as well and can be used to generate high quality images for any of the other 9 passband. Thus, by using GANs we generate AIA0304 images using HMI images when the AIA0304 images are not available to scientists. In this way we can provide scientists and researchers access to complete solar magnetogram and UV/EUV data.

\subsection{Contributions}
In this section the major contributions of this work are highlighted:
 \begin{itemize}
  \item As far as we know, our work is the first attempt to leverage Pix2PixHD algorithm for SDO/HMI to SDO/AIA0304 image-to-image translation and we have shown that Pix2PixHD can indeed be used to solve such problems. The results (Relative Error: -0.045, Pixel to Pixel Pearson correlation coefficient : 0.99, PPE10 or percentage of pixels with relative error less than 10 percent : 0.823 and  Structural Similarity Index Measure: 0.96) indicate that GANs can be extremely useful in generating AIA0304 images from SDO/HMI images. The Pix2PixHD algorithm is able to learn complex features and fine details during training and then reproduce them during testing. 
  \item We have modified the original Pix2PixHD algorithm to solve our task, since the original Pix2PixHD algorithm was designed to synthesize high-resolution images from semantic label maps using conditional generative adversarial networks, while we have employed the Pix2PixHD algorithm for our paired image-to-image translation task. 
  \item We also have modified and implemented the Pix2Pix algorithm to serve as the benchmark model. We have modified the original Pix2Pix algorithm to be able to generate high resolution images. To do this we modified the Generator's neural network architecture. 
  \item Our Pix2PixHD model outperforms the Pix2Pix benchmark model and the previously implemented models by a substantial amount. Our model was trained using only 1892 images, which indicates the effectiveness of our model in small data sets. 
\end{itemize}

\section{Related Work}
Galvez et al.\cite{Galvez_2019} used Convolutional Neural Networks or CNN \cite{Lecun726791} to map  three channel 256 x 256 HMI images to 9 channel 256 x 256 AIA images. Park et al.\cite{Park_2019} applied deep learning for paired image to image translation from solar magnetograms to solar UV and EUV images. To do this they considered two CNN models with two different loss functions one with just L1 loss ($L1$) and the other with L1 plus the cGAN loss ($L_{cGAN}$). Kim et al.\cite{Kim2019} used cGAN\cite{Isola} to generate solar farside magnetograms from STEREO/Extreme UltraViolet Imager  (EUVI) 304-Å images. For this they used pairs of SDO/AIA0304-Å images and SDO/HMI magnetograms to train their deep learning model and then generated solar farside magnetograms using the STEREO/Extreme UltraViolet Imager  (EUVI) 304-Å images. Shin et al.\cite{Shin_2020} used the Pix2PixHD model to generate high resolution magnetograms from Ca II K images. They used pairs of Ca II K 393.3 nm images from the Precision Solar Photometric Telescope at the Rome Observatory\cite{1997jena.confE..37E} and SDO/HMI line-of-sight magnetograms to train their model.
Jeong et al.\cite{Jeong_2020} used pairs of SDO/AIA three passband (304, 193, and, 171) images and the SDO/HMI LOS 720s magnetograms to train their Pix2PixHD model and then generate solar farside magnetograms using the STEREO/EUVI A and B (304, 195, and 171 Å) passband images.

\section{Data and Methodology}
In this section we describe about the Data and the Method used. The Data subsection provides description about the data source, file format, processing and preparation for deep learning application. The Methods subsection includes details about the model's architecture, the loss function, model training and  evaluation framework used.
\subsection{Data}
The datasets used is available on the Joint Science Operations Center (JSOC) server where the Solar Dynamics Observatory (SDO) data can be found. To get the data, perform image processing and prepare the data for the deep learning we used Python packages like SunPy and Astropy. The data are available in Flexible Image Transport System (FITS) file format which is the most extensively used digital file format for storing astronomical data. The FITS file contains metadata which are vital for image processing, visualization and analysis. The AIA0304 and HMI images were extracted from their FITS files respectively. Aligning, rotating, centring and converting to 1024 x 1024 RGB images was carried out. Some of the images with poor quality, excessive noise due to solar flares and obstruction because of planetary transit were manually removed. For model training we are using data for the year 2012, 2013 and 2014 excluding the data for October, November and December of the year 2014 which are used for model testing. This gives us a total of 1892 HMI-AIA0304 image pairs for model training and 200 image pairs for testing and evaluation. We did not include the data for the subsequent years due to hardware constraints and due to the subsequent degradation of the SDO/AIA0304-Å instrument which results in loss of details captured in the images.

\subsection{Methods}
The following section describes the Pix2Pix and the Pix2PixHD neural network architecture along with the objective or the loss function used, the model training process and the model evaluation metrics used to quantify the model's performance.
\subsubsection{Pix2Pix Network Architecture, Objective Function and Model Training}
\paragraph{Network Architecture:}
The algorithm used is the conditional image-to-image translation Generative Adversarial Network, cGAN\cite{Isola}. Since the original Pix2Pix algorithm was not designed to generate high resolution images, we have modified the original algorithm to handle 1024x1024, 3 channel RGB solar images. The modifications include adding additional layers to the U-Net\cite{10.1007/978-3-319-24574-4_28} architecture which is described below. The GAN consists of a Generator which has a U-Net \cite{10.1007/978-3-319-24574-4_28} architecture and the Discriminator is a PatchGAN \cite{Isola} classifier. To handle 1024x1024 images, in our algorithm the Generator contains 10 encoder and decoder blocks respectively. Each encoder blocks consisting of Convolutional layers, Batch Normalisation and LeakyReLU activations while each decoder blocks is consisting of Transposed convolutions, Batch Normalisation, ReLU activations. The U-Net also contains skip connections that concatenate the encoder and the decoder blocks at the same resolution to prevent loss of important information during the forward pass for the encoding stage and also help in improving the flow of gradients back to the encoder during the backward pass, preventing the problem of vanishing gradients. Dropout layers are also used in some blocks to incorporate stochasticity into the network during training. The Discriminator is a PatchGAN. A PatchGAN is a CNN that tries to classify patches of an image as real or fake, rather than the whole image. The output of a PatchGAN a matrix (2d-array) instead of just a single value. PatchGAN consists of Convolutional layers, Batch Normalizations and LeakyReLU activations. 

\paragraph{Objective function:}The objective or the loss function for the Discriminator is the traditional adversarial loss function, that is the negative log likelihood minimization for recognizing fake and ground truth images conditioned on the ground truth image. The Generator on the other hand is trained using the adversarial loss along with the $L_1$ or pixel distance loss between the generated image and the real or target image. The $L_1$ loss encourages the generated image for a particular input to remain as similar as possible to the corresponding output real or ground truth image. This leads to faster convergence and more stable training. The loss function for conditional GAN is given by

\begin{equation}\label{Conditional GAN loss}
L_{cGAN}(g,\ d)\ =E_{x,y}[log\ d(x,\ y)]+ E_x[log(1 - d(x,g(x))],
\end{equation}

where the generator g tries to minimize the above function whereas the discriminator d tries to maximize it. In the above equation $x$ is the input, $y$ is ground truth, $g(x)$ is the generated fake output. The above equation can also be written as
\begin{equation}
    g^{\ast\ }={min}_g{max}_dL_{cGAN}(g,\ d)\     
\end{equation}

The $L1$ loss or the pixel distance loss is given by
\begin{equation}
   L_1=E_{x,y}({|\left|y-g\left(x\right)\right||}_1) 
\end{equation}

The goal of the generator is to minimize the above pixel distance loss term hence the final loss/objective can be written as 
\begin{equation}
    g^\ast={min}_g{max}_dL_{cGAN}\left(g,d\right)+\lambda\ L_1(g),
\end{equation}

where $\lambda$ is the weighting hyper-parameter coefficient. In our case we took $\lambda$=100 similar to the what was used by Isola et al.\cite{Isola}.

\paragraph{Pix2Pix Model Training:}
During training the goal of the Generator which has a U-Net architecture is to take in as input SDO/HMI images and then generate images that look like the real SDO/AIA0304. The generator tries to minimize the conditional GAN loss ($L_cGAN$) along with the pixel distance ($L_1$) loss. The discriminator on the other hand tries to determine which ones are the real pairs of SDO/AIA0304, SDO/HMI images and which are the generated(fake) pairs by minimizing the adversarial loss. Figure 1. shows the working of the Pix2Pix image-to-image translation conditional GAN model.

\begin{figure}[h!]
    \centering
    \includegraphics[width=\textwidth]{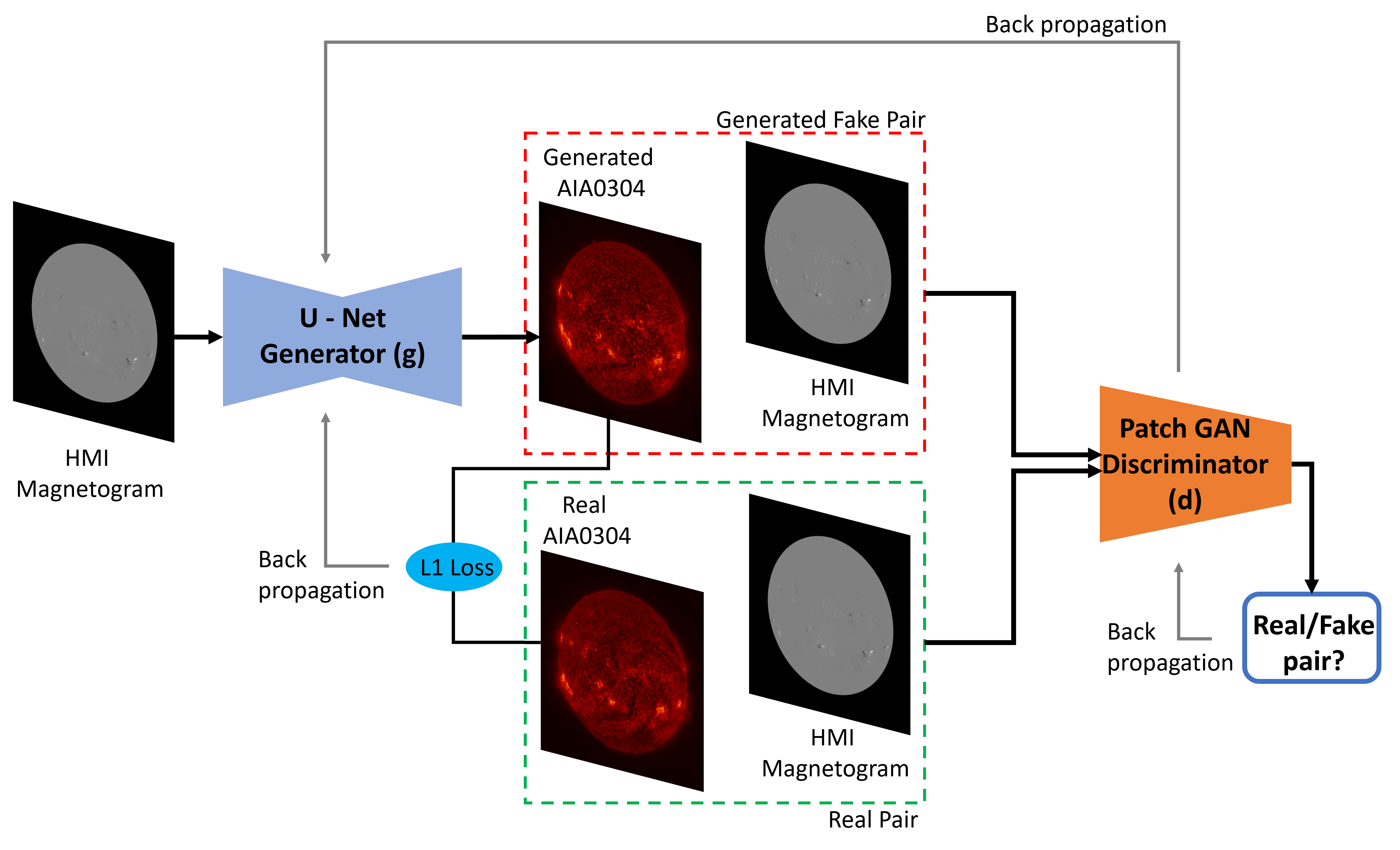}
    \caption{\textbf{Pix2Pix Conditional GAN model for paired image-to-image translation}}
    \label{fig:gan model}
\end{figure}

Both the real and the generated AIA0304 images are concatenated with SDO/HMI images respectively along the channel dimension to form the real and generated pairs of images. The Discriminator which is a PatchGAN, attempts to identify which one is the real pair and which one is the fake/generated pair by outputting a matrix of values between 0 and 1 which shows how real of how fake different parts of the image are. Our model was trained for 200 epochs on Google Colab Pro with GPU acceleration. For every 10 epochs the generator model was saved for making predictions after model training. This gives us a total of 20 generator models from which we can choose the best model during model evaluation on the unseen test data.  

\subsubsection{Pix2PixHD Network Architecture, Objective Function and Model Training}

\paragraph{Network Architecture:} The Pix2PixHD\cite{WangPix2PixHD} is an improved version of the Pix2Pix algorithm. The primary goal of Pix2PixHD is to produce high-resolution images. To do this the authors introduced multi-scale generators and discriminators. Pix2PixHD conditional GAN consists of a Coarse to Fine Generator which has two components first the Global generator G1 and the Local enhancer G2. There is no difference between G1 and the Pix2Pix generator, it is an end2end U-Net structure. The Global generator G1 is made up of 3 components that include a convolutional front-end (encoder), a set of residual blocks and a transposed convolutional back-end (decoder). The local enhancer is made up of 3 components which include a convolutional front-end , a set of residual blocks and a transposed convolutional back-end. The global generator is sandwiched between two halves of the G2 generator. The left half of G2 extracts features, fuses them with the previous layer features of G1's output layer, and sends the resulting information to the right half of G2 to produce a high-resolution image. The network architecture of the discriminator remains unchanged from the Pix2Pix algorithm, i.e. a patch based fully connected convolutional neural network. What changes is that we no longer just use one discriminator instead we use multiple (in our case two) discriminators that operate at different scales. Having two discriminators operating at different scales encourages the generator to produce images that are consistent globally without lacking the local finer details.

\paragraph{Objective function:} The authors of Pix2PixHD improved upon the traditional conditional GAN loss by incorporating the Feature matching loss $L_{FM}$ based on the discriminator. We have already described the $L_{cGAN}$ loss in the previous section. The feature matching loss acts as regularization and stabilizes training as the generator now has to produce similar statistics at multiple scales. The feature matching loss is given by

\begin{equation}\label{feature matching loss}
L_{FM}(g,\ d)\ =E_{x,y}\sum_{i=1}^{T}[
  \left\lVert d^{i}(x,y) -  d^{i}(x,g(x) \right\rVert_1],
\end{equation}
where $T$ is the total number of layers, $i$ is the serial number of the layers and $N_i$ represents the number of pixels in each layer. The objective function is given by
\begin{equation}
    {min}_g(({max}_{d_1,d_2}\sum_{k=1,2}L_{cGAN}(g,d_k))) + \lambda \sum_{k=1,2} L_{FM}(g,d_k))),
\end{equation}
where $k$ denotes the serial number of the two discriminator and $\lambda$ denotes the weighting hyperparameter. We use the value of $\lambda$ to be 10 which was used by Wang et al.\cite{WangPix2PixHD}.

\paragraph{Pix2PixHD Model Training:}Similar to the training of the Pix2Pix algorithm described above. Figure 2. shows the working of the Pix2PixHD image-to-image translation conditional GAN model. During training of the Pix2PixHD model, the input is the HMI image and the output is the generated AIA304 iamge. The network then updated the parameters based on the calculated values of the $L_{cGAN}$ and the $L_{FM}$. We train the model for 200 epochs and saved the generator model every 10 epochs for testing on the test data. 

\subsubsection{Model Evaluation}
To quantify the model’s performance and compare the real SDO/AIA0304 images with the GAN generated ones for the entire test dataset we used the following four types of evaluation metrics. 

\begin{figure}[h!]
    \centering
    \includegraphics[width=\textwidth]{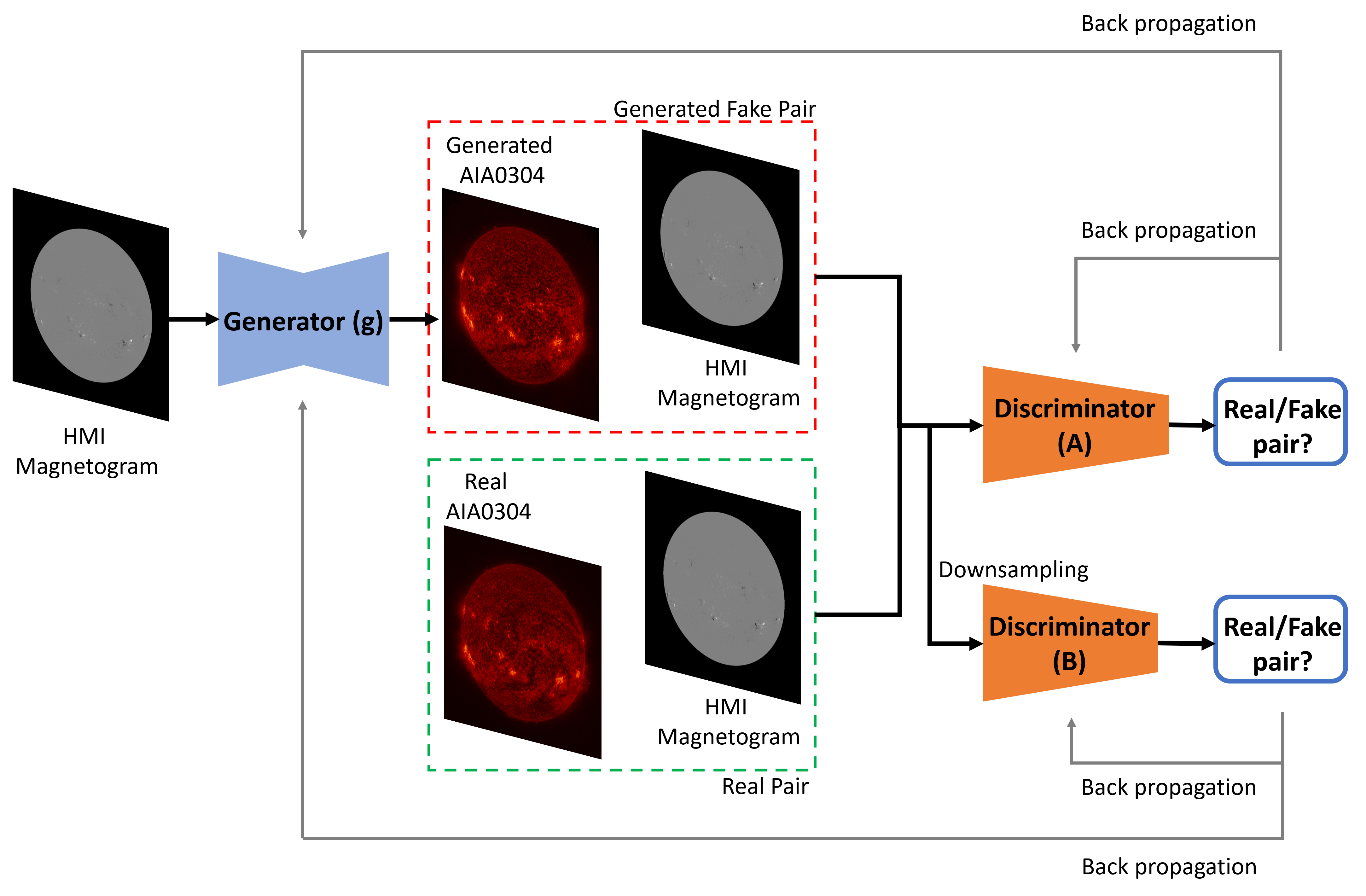}
    \caption{\textbf{Pix2PixHD GAN model for paired image-to-image translation}}
    \label{fig:gan model}
\end{figure}

\begin{itemize}
 \item The first model evaluation metric is the Relative Error (RE). Relative error of the total pixel value ${(\Phi}_i)$ is given by
 \begin{equation}
    {RE}_i=\frac{abs(\Phi_i^{GAN}-\Phi_i^{AIA0304})}{\Phi_i^{AIA0304}},
 \end{equation}
 where $i$ represents the serial number of the 200 testing image data samples. Lower absolute value of relative indicates better performance. 

 \item The second metric for evaluation is the Pixel to Pixel Pearson Correlation Coefficient(CC). A higher value of CC denotes that the GAN model is not only able to generate the correct pixel values but also the values are spatially correct. 

 \item The third metric is the Percentage of Pixel Having Relative Error Less than 10\% (PPE10) which finds the fraction of the number of pixels in an image which have relative error less than 10\% and indicates the percentage of good pixels in the generated image. Higher PPE10 value indicates better performance. 

 \item The final model evaluation metric is the Structural similarity index measure (SSIM) \cite{Wang} which is a method for quantifying the similarity between two images. SSIM tries to model the perceived change in the image's structural information. The SSIM value varies between -1 and 1, where a value of 1 shows perfect similarity and a higher value indicates better performance.
\end{itemize}

\begin{figure}[h!]
\centering
    \begin{subfigure}{1.0\linewidth}
    \centering
    \includegraphics[width=\textwidth]{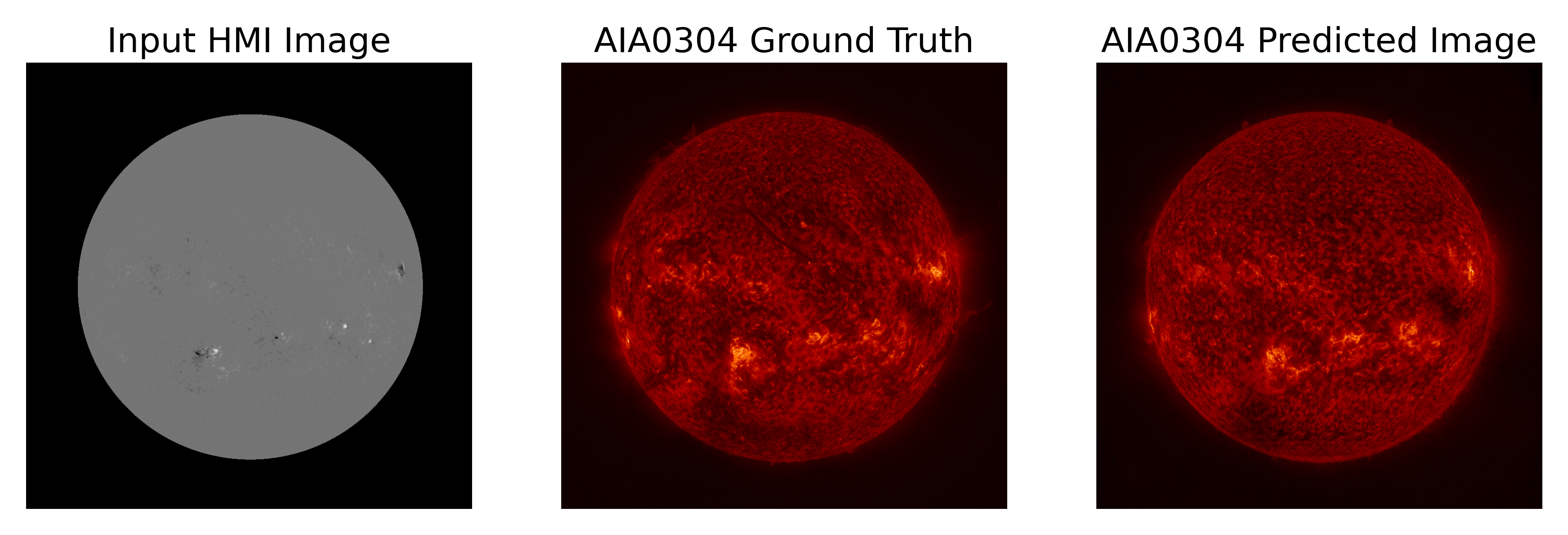}
    \caption{\textbf{September 01, 2014}}
    \label{fig:image1}
    \end{subfigure} %

    \hfill

    \begin{subfigure}{1.0\linewidth}
    \centering
    \includegraphics[width=\textwidth]{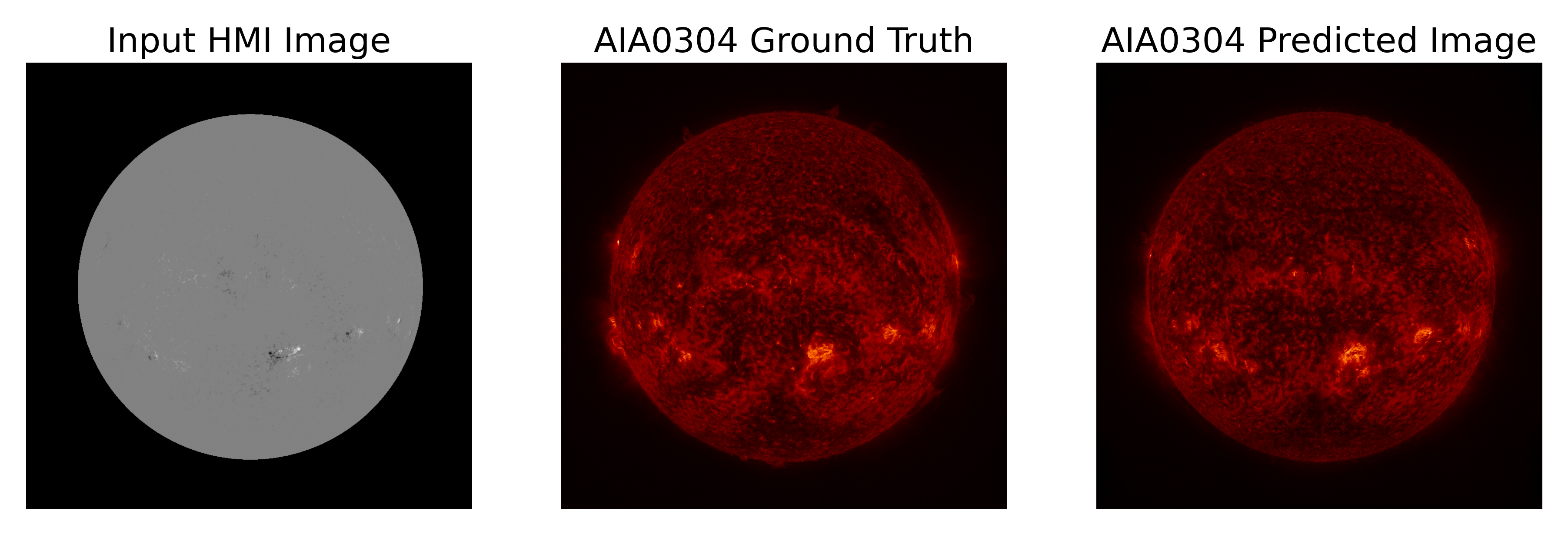}
    \caption{\textbf{September 05, 2014}}
    \label{fig:image2}
   \end{subfigure}

   \hfill
   
   \begin{subfigure}{1.0\linewidth}
    \centering
    \includegraphics[width=\textwidth]{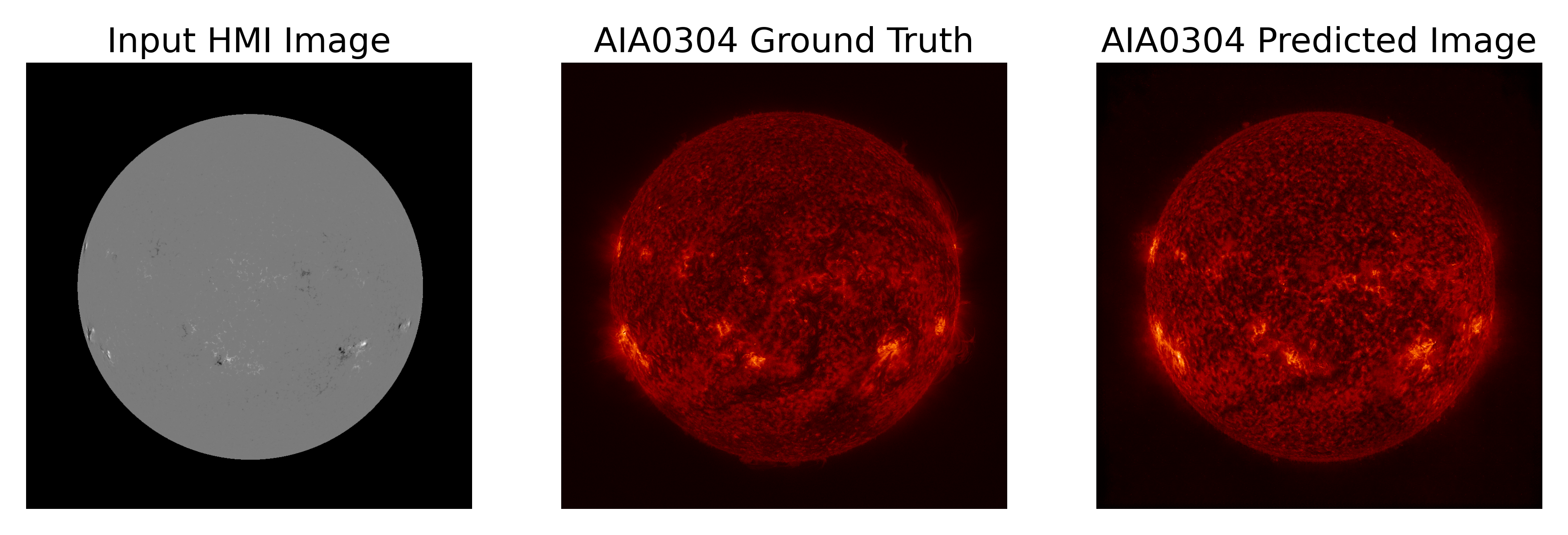}
    \caption{\textbf{September 09, 2014}}
    \label{fig:image3}
   \end{subfigure}

\caption{\textbf{Comparison between real AIA0304 and Pix2PixHD cGAN generated ones.} (a), (b) and (c) of Figure 4 show the input HMI image, the ground truth AIA0304 image and the GAN generated AIA0304 images respectively taken for the month of September with 4 day cadence.}
\label{fig:output}
\end{figure}

\section{Results and Discussion}
From Figure 3. we can see the SDO/HMI magnetogram images as input along with the ground truth SDO/AIA0304 images and the Pix2PixHD generated AIA0304 images. In the figure we can observe that the Pix2PixHD model learns the active region and quiet region patterns for the full disc during training and then is able to generate these regions during testing/image generation step. The Pix2PixHD generated images are of high quality and are comparable to the real images. 

Table 1. presents the evaluation metric values for the entire test dataset consisting of 200 images for the month of September, October, November and December for the year 2014. We compare the results of the our Pix2PixHD model and Pix2Pix model with the baseline model trained using the algorithm developed by Park et al. \cite{Park_2019}.

\begin{table}[h!]
    \centering
\begin{tabular}{|p{0.4\linewidth}|p{0.2\linewidth}|p{0.15\linewidth}|p{0.15\linewidth}|}
    \hline
\textbf{Metrics}    & \multicolumn{3}{c|}{\textbf{Full disk 200 Images}}
                          \\
  &   Pix2PixHD (ours) & Pix2Pix (ours)  & Park et al.\cite{Park_2019} \\
    \hline
Relative Error (RE) & -0.045 & 0.055 & 0.061  \\
    \hline
Pixel-to-pixel Pearson correlation coefficient (PCC) & 0.99 &  0.962 & 0.933 \\
    \hline
Percentage of Pixel Having Error Less than 10\% (PPE10)	& 0.823 & 0.681 & 0.652 \\
    \hline
Structural Similarity Index Measure (SSIM)	& 0.96 & 0.884 & 0.851\\
    \hline
\end{tabular}
    \caption{\textbf{Model Evaluation metrics for Pix2PixHD and Pix2Pix-generated images for the full disk. Comparing results with Park et al.\cite{Park_2019}}}
    \label{Table1}
\end{table}

From Table 1. we can observe that our Pix2Pix model and Pix2PixHD model are able to get better results than the results obtained by utilizing the model trained using Park et al.\cite{Park_2019} algorithm on our test dataset. The Pix2PixHD model outperforms all the other models. The relative error of -0.045 suggests that the Pix2PixHD model Generator underestimates the pixel values slightly while the  Pix2Pix models overestimate the the pixel values. The Pixel-to-pixel Pearson correlation co-efficient (PCC) value of 0.99 for Pix2PixHD indicates strong linear relation between the generated and the real images. The other algorithm used by Park et al.\cite{Park_2019} performs slightly less than our Pix2Pix model. The Pix2PixHD algorithm also does much better than the other algorithms with more than 14 percent increase for the PPE10 values, this shows the Pix2PixHD model makes less than 10 percent error on much higher number of pixels when compared to other models. Finally the SSIM value of 0.96 for the Pix2PixHD model demonstrates that it is able to learn and generate intricate structures in the images. The algorithm used by Park et al.\cite{Park_2019} is also based upon the Pix2Pix algorithm therefore the performance is similar though slightly lower than our Pix2Pix model which indicate that the modifications we made to the original Pix2Pix algorithm helps our model to achieve slight improvement over previously implemented models. Thus by implementing the Pix2Pix model we prove that the  Pix2Pix, which is the benchmark image to image translation algorithm is also able to learn the structures of the images and obtain reasonable values.  However the Pix2PixHD model clearly surpasses our Pix2Pix model and all other previous works  in generating high-resolution images. This can be attributed to the fact that instead of using single Generator and Discriminator architecture similar the Pix2Pix algorithm, Pix2PixHD algorithm uses a multi-scale network structure for the Generator and the Discriminator which helps to stabilize training, generate high resolution and vivid images. Also, the addition of the feature matching loss helps to stabilize model training, improve image resolution, color and texture.

The above results show that the Pix2PixHD model can successfully generate SDO/AIA0304 images from SDO/HMI magnetograms. Note that our Generative Adversarial model was trained using just 1892 images the results that we got are promising. Typically other deep learning algorithms such as CNNs generally require copious amounts of data to identify patterns in the images and solve similar image-to-image translation problems.

\section{Conclusion and Future Work}
Here we used the conditional Generative Adversarial Network framework (Pix2PixHD) to perform image-to-image translation, from SDO/HMI images to SDO/AIA0304 images. As far as we know, this is the first time the Pix2PixHD algorithm has been used for SDO/HMI to SDO/AIA0304 image-to image- translation and we have shown that Pix2PixHD can indeed be used to solve such problems. The results (Relative Error: -0.045, Pixel to Pixel Pearson correlation coefficient : 0.99, PPE10 or percentage of pixels with relative error less than 10 percent : 0.823 and  Structural Similarity Index Measure: 0.96) indicate that GANs can be extremely useful in generating AIA0304 images from SDO/HMI images. The Pix2PixHD algorithm is able to learn complex features and fine details during training and then reproduce them during testing. We used data for the year 2012, 2013 and 2014 for model training excluding the data for October, November and December of the year 2014 which were used for model testing. The choice for the amount of data used was governed by hardware constraints and degradation of the SDO/AIA0304 instrument. In the future iteration we plan to use the data for the subsequent years by applying degradation correction factor to account for the degradation of AIA0304-Å instrument. Since the data are temporally and spatially oriented the use of Video-to-Video Synthesis models like Vid2Vid can also be explored.

\printbibliography

\end{document}